%% file: 0CAMERAREADY.tex
\pdfoutput=1

\documentclass[11pt]{article}

\usepackage{acl}

\usepackage{times}
\usepackage{multirow}
\usepackage{booktabs} 
\usepackage{latexsym}
\usepackage{enumitem}
\usepackage{algorithm, tabularx}
\usepackage[noend]{algpseudocode}
\usepackage[title]{appendix}
\usepackage{listings}
\usepackage{float}
\usepackage{hyperref}
\usepackage{adjustbox}

\usepackage[T1]{fontenc}

\usepackage[utf8]{inputenc}

\usepackage{graphicx}
\usepackage{booktabs}

\usepackage{microtype}
\usepackage{csquotes}

\usepackage{amsmath}

\makeatletter
\newcommand{\multiline}[1]{%
  \begin{tabularx}{\dimexpr\linewidth-\ALG@thistlm}[t]{@{}X@{}}
    #1
  \end{tabularx}
}
\makeatother

\newcommand{\llama}{\includegraphics[height=1em]{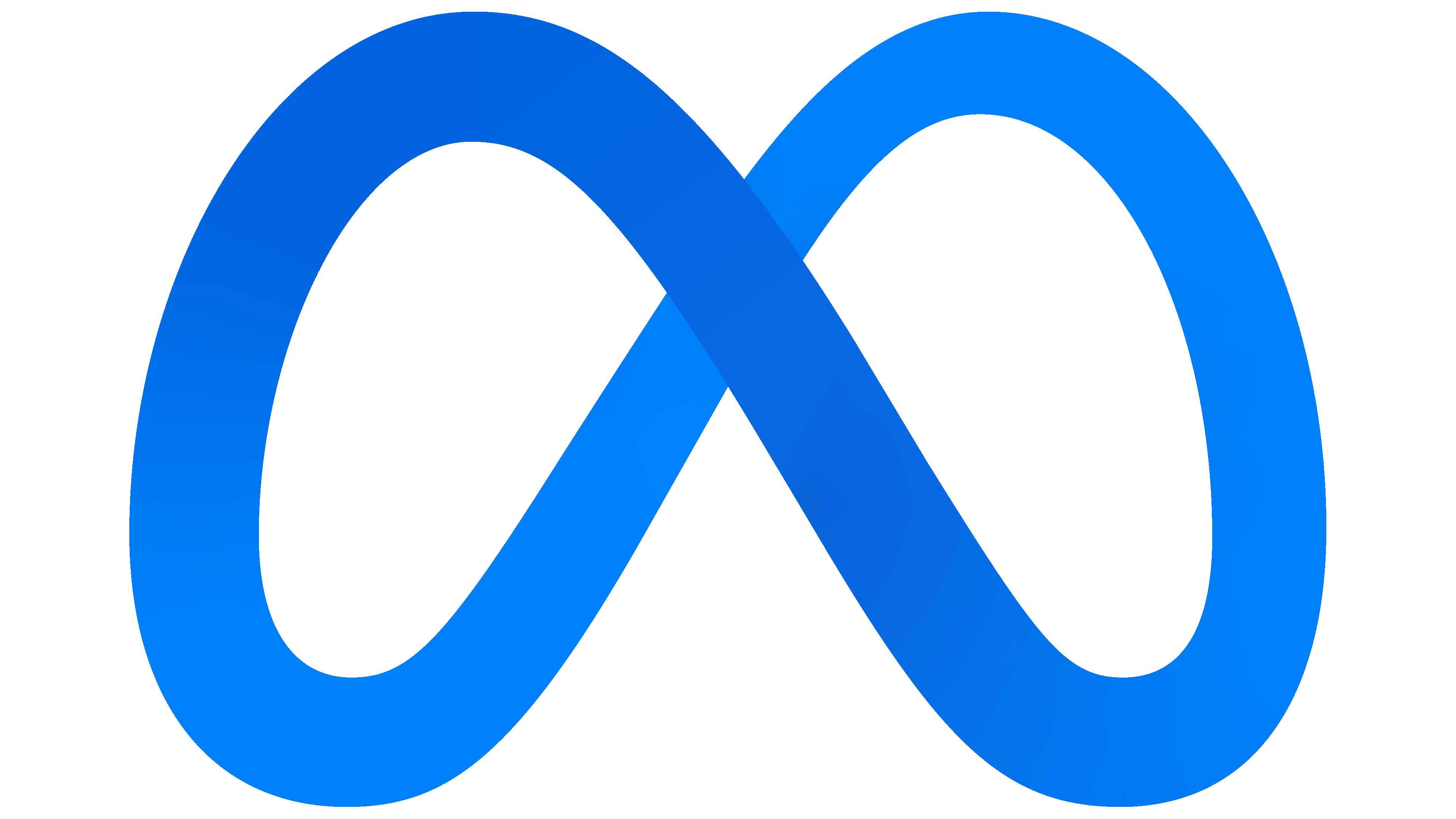}}
\newcommand{\gpt}{\includegraphics[height=1em]{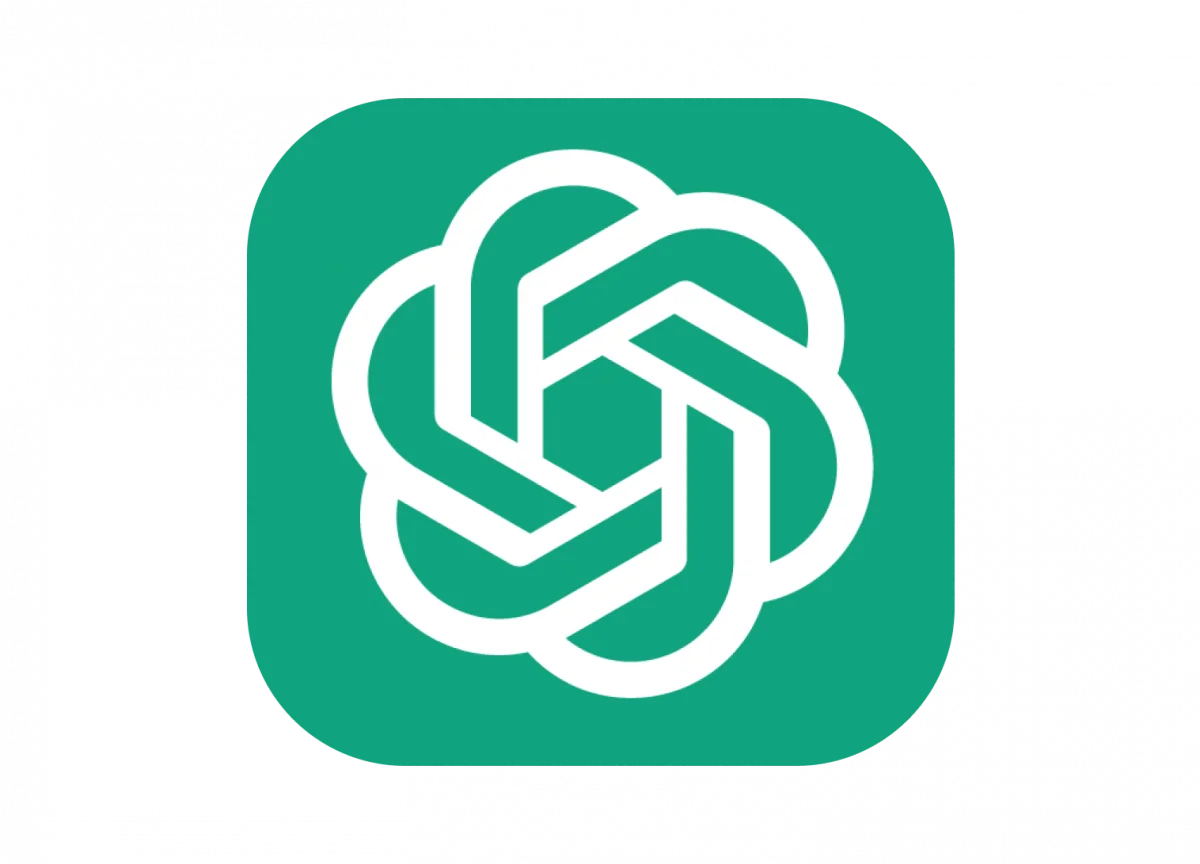}}

\title{Instruct, Not Assist: \\LLM-based Multi-Turn Planning and Hierarchical Questioning \\for Socratic Code Debugging}

  \author{Priyanka Kargupta$^*$, Ishika Agarwal$^*$, Dilek Hakkani-Tur, Jiawei Han\\
  $^{1}$Department of Computer Science, University of Illinois at Urbana-Champaign\\
  \texttt{\{pk36, ishikaa2, dilek, hanj\}@illinois.edu} \\
}

\begin{document}
\maketitle

\def\thefootnote{*}\footnotetext{These authors contributed equally to this work.}\def\thefootnote{\arabic{footnote}}

\begin{abstract}

Socratic questioning is an effective teaching strategy, encouraging critical thinking and problem-solving. The conversational capabilities of large language models (LLMs) show great potential for providing scalable, real-time student guidance. However, current LLMs often give away solutions directly, making them ineffective instructors. We tackle this issue in the code debugging domain with \textbf{TreeInstruct}, an Instructor agent guided by a novel state space-based planning algorithm. TreeInstruct asks probing questions to help students independently identify and resolve errors. It estimates a student's conceptual and syntactical knowledge to dynamically construct a question tree based on their responses and current knowledge state, effectively addressing both independent and dependent mistakes concurrently in a multi-turn interaction setting. In addition to using an existing single-bug debugging benchmark, we construct a more challenging multi-bug dataset of 150 coding problems, incorrect solutions, and bug fixes-- all carefully constructed and annotated by experts. Extensive evaluation shows TreeInstruct's state-of-the-art performance on both datasets, proving it to be a more effective instructor than baselines. Furthermore, a real-world case study with five students of varying skill levels further demonstrates TreeInstruct's ability to guide students to debug their code efficiently with minimal turns and highly Socratic questioning.
\end{abstract}

\begin{figure}[t]
    \centering
\includegraphics[width=0.48\textwidth]{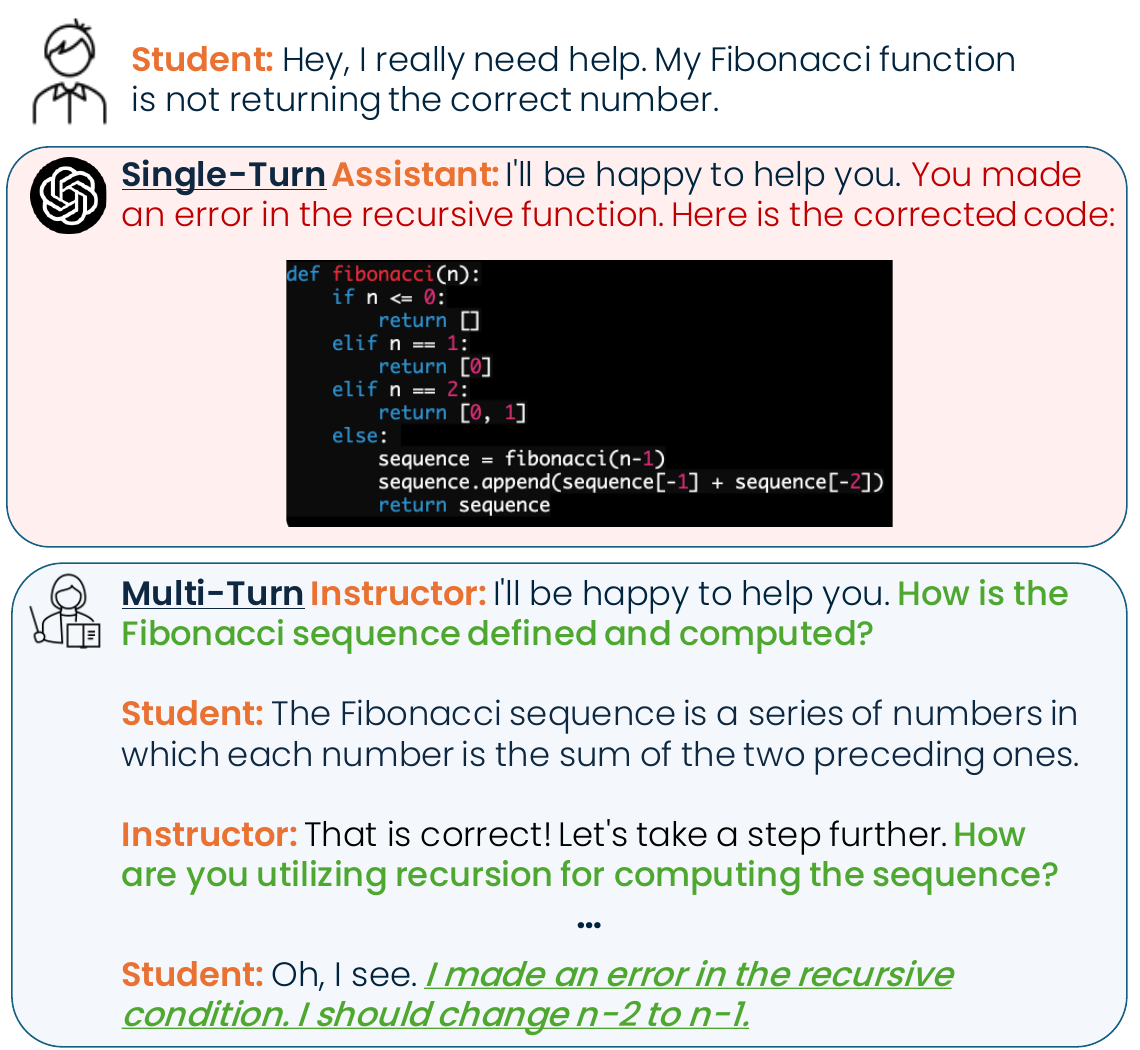}
    \caption{The Instructor's goal is to generate multi-turn Socratic questions while guiding the Student towards the correct solution.}
    \label{fig:case_sg}
    \vspace{-0.3cm}
\end{figure}

\section{Introduction}

\par With the rapidly expanding conversational and reasoning abilities of large language models (LLMs), there has been a substantial rise in demand for exploiting their capabilities within a multitude of educational applications \cite{kasneci2023chatgpt} in order to widen accessibility via personalized feedback. Specifically, several recent works explore the use of LLMs for providing feedback and guidance to students \cite{wang2023bridging, kazemitabaar2024codeaid, sheese2024patterns, lyu2024evaluating}. However, LLMs are typically optimized to generate customer-serving, assistant-like responses, which also translates into the types of questions asked. Especially for educational domains, this style of questioning can be suboptimal \cite{cotton1988classroom, sahamid2016developing, yang2005using, wilson1987socratic}. For instance, if a student is seeking help from an instructor for correcting their mistakes (e.g., debugging their buggy code), we consider two forms of potential responses: \textbf{assistant-like} and \textbf{instructor-like}. As shown in Figure \ref{fig:case_sg}, an assistant-like response would not be a successful educational interaction, as it leads to the Assistant directly providing an answer. On the other hand, an Instructor-like response reflects the educational philosophy of \textit{Socratic questioning}.

\par Socratic questioning is a teaching strategy where the Student independently solves their problem by answering \textit{guiding} questions, instead of being given the \textit{solution directly} \cite{wilson1987socratic}. This is a more effective learning strategy because the weight of learning falls on the Student as they must put in effort to answer a question as opposed to solely relying on the model \cite{cotton1988classroom, kasneci2023chatgpt}. Therefore, we aim to re-orient an LLM to be an Instructor, not an assistant, by asking Socratic questions that (1) help the Student understand their mistakes, and (2) do not directly provide the answer. To tackle these challenges, we propose \textbf{TreeInstruct} based on the following principles:
\begin{enumerate}[leftmargin=*]
\itemsep-0.15em 
    \item \textbf{State space estimation:} An Instructor plans its conversation with a Student based on the ``distance'' between their initial answer and the optimal, correct answer within the estimated state space. In other words, it tracks the knowledge state of the Student within this space throughout the Instructor-Student interactions.
    \item \textbf{Tree-based Socratic questioning:} An Instructor generates turn-level Socratic questions conditioned on both the Student's current knowledge state \textit{and} misunderstanding(s), the latter derived from their responses to the Instructor's questions. This step dynamically constructs a Socratic question tree.
    \item \textbf{Adaptive conversation restructuring:} An Instructor updates their initial conversation plan based on how the Student is progressing in the conversation, as reflected by updates (or lack thereof) to the Student's knowledge state. This planning can include both \textit{questioning} and \textit{teaching} actions.
\end{enumerate}

\par While these principles can apply to many educational domains, this paper focuses on code debugging, which presents unique challenges. Real-world code debugging often involves multiple, potentially interdependent conceptual and syntactical bugs. For instance, Figure \ref{fig:case_sg} shows that first resolving the Student's conceptual misunderstanding of recursion in Fibonacci helps them identify their recursive syntactical bug (Figure \ref{fig:case_sg}). However, existing work fails to account for such nuances and assumes single-turn feedback \cite{kazemitabaar2024codeaid, wang2023bridging, lyu2024evaluating}. This ignores the sub-steps required for the Student to understand each bug.

\par In contrast, TreeInstruct constructs a multi-turn debugging plan (\textit{state representation}), defined as the set of Student misunderstandings and mistakes (\textit{state variables}) to be resolved in order to comprehend and correct their bug(s). We define all potential paths to complete these tasks as the \textit{state space}. We traverse the space using Socratic questions and trace which variables have been resolved, grounded based on the Student's responses.

\par While existing LLM-based tutors are effective in fixing the Student's code with high success, they are either prone to directly revealing code answers or cannot be adapted to new Student responses. For example, CodeAid~\cite{kazemitabaar2024codeaid} (specifically, the "Help Fix Code" and "Question from Code" modules, as these are most similar to our setting) directly provides code or pseudocode 57\% of the time, and achieves a mere 55\% rate of helpfulness. On the other hand, TreeInstruct exploits the state space to dynamically construct a tree of questions based on (1) incorrect Student responses, or (2) gaps in the Student's knowledge. The sibling and parent-child relationships between questions reflect the manner in which they traverse the state space. Finally, it exploits both the Student's knowledge state and any proposed bug fixes to serve as the dynamic stopping condition. Overall, TreeInstruct takes a more structured approach to multi-turn conversational feedback, as (1) grounding the conversation on the state space representation ensures that all bugs are sufficiently addressed, and (2) constructing a tree based on the Student's current level of understanding allows for more relevant and personalized question generation.


\noindent We summarize our contributions below:
\begin{itemize}[leftmargin=*]
\itemsep-0.35em 
    \item To the best of our knowledge, TreeInstruct is the first work to explore state space estimation and dynamic tree-based questioning for multi-turn Socratic instruction. 
    \item We construct a novel multi-bug debugging dataset with 150 expert-annotated, challenging conceptual and syntactical bugs and their fixes. 
    \item Extensive experiments on an existing benchmark and our constructed dataset demonstrate that TreeInstruct can be universally applied to both open and closed source-settings. 
    We also showcase that TreeInstruct's strong Socratic questioning abilities widely outperform all baselines through both (1) rigorous quantitative and qualitative expert evaluation (on average, \textit{preferred} 78.43\% of the time; Student fixes code 24.55\% \textit{more}) and (2) real-world interactions with students of varying coding abilities.
\end{itemize}
\noindent
\textbf{Reproducibility:} We release our data and source code\footnote{https://github.com/agarwalishika/TreeInstruct} to facilitate further studies.
 
\input{algorithm}
\section{Related Works}
\label{section: related_work}
\subsection{Knowledge Tracing}
Knowledge tracing tracks student knowledge to personalize their learning experience, including understanding specific concepts, behavior, and recall ability. There are two primary methods: probabilistic and deep learning-based.
Probabilistic knowledge tracing, as it was first introduced, uses a Hidden Markov Model (HMM) to maintain binary states, learned and unlearned, for each skill as learners engage with exercises. This approach, from which we draw inspiration, updates the likelihood of these states based on performance \cite{corbett1994knowledge, bkt}. Some models use open-ended paths to states \cite{rafferty2016faster}, while others use deep learning-based, long-term memory capabilities essential for learning \cite{piech2015deep}. These methods are performative, but such state spaces hinder effectiveness and require large amounts of annotated training data.

Our methodology addresses the challenge of limited annotated data by dynamically generating states during interactions between instructors and students. We monitor these evolving states through a component we refer to as the Verifier. Using these dynamically generated states, we tailor the educational experience by personalizing the sequence and type of questions posed to learners.

\subsection{Socratic Reasoning in Educational AI} 
There have been several works exploring Socratic reasoning in education \cite{herbel2005questioning, wang2024convokit, alic2022computationally, demszky2022ncte}. More recently, prior work ~\cite{al2023socratic, al2023can} has highlighted the poor performance of prompting-based methods in performing Socratic Reasoning for the education domain ~\cite{achiam2023gpt}, even with Chain-of-Thought (CoT)~\cite{wei2022chain}, as they often give away answers without asking clarifying questions, or the questions are unrelated to the student's response or original bug \cite{achiam2023gpt}. In contrast, TreeInstruct mitigates this issue by explicitly grounding the question generation step on both a target state variable $\tau$ and any Student \textit{misunderstanding} gauged from their previous response.

\subsection{LLMs for Interactive Education}

Recent generative approaches within the AI tutoring space have attempted to generate responses which cater to the student's type of mistake or request, but only in single-turn settings. CodeAid \cite{kazemitabaar2024codeaid} is an \textit{assistive} tool that helps students debug their code. In their "Help Fix Code" and "Question from Code" modules, the Instructor provides single-turn responses to the Student for answering questions, explaining concepts, and helping to write code. However, these modules direct the Student towards where their mistake is and uses natural language to describe the bug fixes. In contrast, TreeInstruct aims to \textit{instruct} the Student socratically through questions, such that even in natural language, the bug fixes are not provided. BRIDGE \cite{wang2023bridging} is an Instructor-like framework that aims to help students with math mistakes. The LM estimates the type of error, the strategy of error remediation, and the instructor intention behind the remediation (all are chosen from a predetermined set). However, our methodology makes use of a more structured planning approach that accounts for the inherent multi-turn nature of educational guidance.

\begin{figure}[h]
    \centering
    \includegraphics[width=0.5\textwidth]{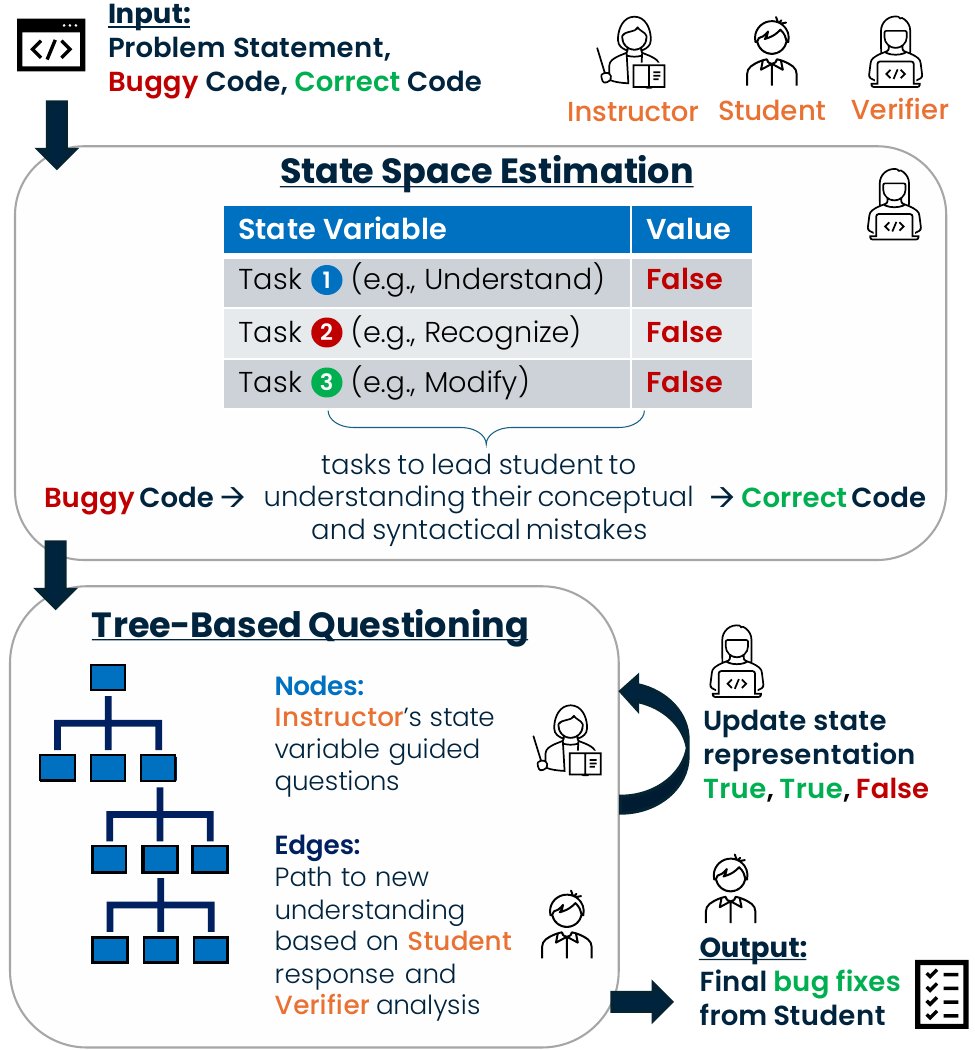}
    \caption{
    \label{fig:framework} We propose TreeInstruct, a novel tree-guided instructional questioning framework for meaningful educational debugging guidance.}
    \vspace{-0.3cm}
\end{figure}

\begin{figure*}[ht]
    \centering
    \includegraphics[width=1.0\textwidth]{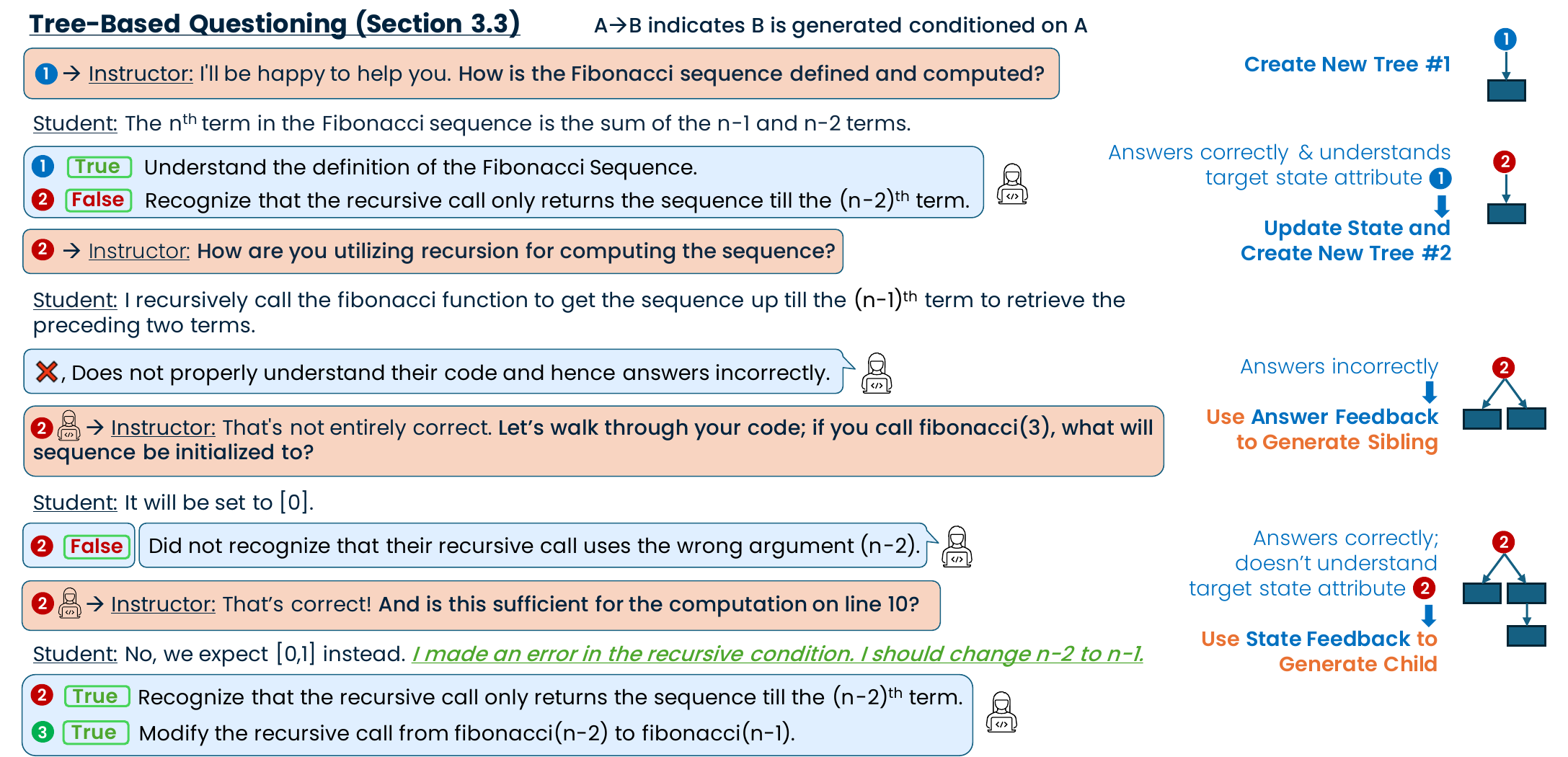}
    \caption{
    \label{fig:tree} We detail the process for tree-based question generation. Blue and orange text/backgrounds indicate that the Instructor and Verifier are performing the task, respectively.}
\end{figure*}

\section{Methodology}



\par As shown in Figure \ref{fig:framework}, \textbf{TreeInstruct} aims to dynamically guide the multi-turn conversation based on its estimated state space. Section \ref{sec: agents} provides an overview of the three different agents we use and their respective roles during the state space generation/update and tree construction processes (outlined in Figure \ref{fig:tree}). This allows TreeInstruct to respond to the Student's current level of understanding adequately. Algorithm \ref{algorithm: pseudocode} contains the pseudocode for all components in our method.

\subsection{Preliminaries}
\subsubsection{Problem Description}

As input, the Instructor is given the Student's buggy code that contains $e$ errors, a problem statement, bug descriptions, and their respective fixes. The Instructor guides the Student to generate a list of all bug fixes based on their interactions with the Instructor. The overall goal is for the Student to resolve their own conceptual and syntactical errors in a Socratic fashion to reach the correct code. Note that we assume bug fixes are provided, a common scenario in educational settings (e.g., assignments, exams) where ground-truth solutions are available from human instructors or platforms like LeetCode. As our focus is steering an LLM towards Socratic guidance, generating these solutions for real-world tutoring applications is left for future work.

\subsubsection{Agents}
\label{sec: agents}
In a real-world setting, a Socratic educator (e.g., an instructor, a teaching assistant) executes two tasks when interacting with a Student: (1) ask relevant questions to the Student, and (2) assess the Student's understanding based on their responses. Following this cyclical pattern, we break down our educator into two roles: an Instructor and a Verifier, with persona prompts specified in Tables \ref{prompt: instructor_persona} and \ref{prompt: verifier_persona} in Appendix \ref{appendix: prompts}, respectively. The Instructor and Verifier perform their respective tasks specified in Algorithm \ref{algorithm: pseudocode} via zero and one-shot prompting. The \textit{Instructor} agent's job is to generate questions to ask the Student (\hyperref[prompt: i2i_generate_initial]{GenerateQuestion}, \hyperref[prompt: i2i_generate_sibling]{GenerateSiblingQuestion}, and \hyperref[prompt: i2i_generate_child]{GenerateChildQuestion} in Alg. \ref{algorithm: pseudocode}; details provided in Section \ref{sec: tree-based questioning}). The \textit{Verifier} agent has a significantly more involved role:
\begin{enumerate}[leftmargin=*]
    \item \textit{State Space Estimation (Section \ref{sec: state space representation}):} The Verifier determines a set of tasks which will lead a Student to understanding and correcting their problem and buggy code. This is \hyperref[prompt: v2v_get_state_repr]{\texttt{GenerateState}} in Alg. \ref{algorithm: pseudocode}.
    \item \textit{Assess Student Response (Section \ref{sec: tree-based questioning}):} Once the Student answers the Instructor's question, the Verifier must judge the response's accuracy, given the question-answer pair interaction. This is \hyperref[prompt: i2i_correct_response]{\texttt{VerifyResponse}} in Alg. \ref{algorithm: pseudocode}.
    \item \textit{Assess Student Understanding of Target State Variable (Section \ref{sec: tree-based questioning}):} To update the Student's state space representation, the Verifier must determine whether the Student would have needed a sufficient understanding of the target state variable in order to generate their response. This is \hyperref[prompt: i2i_address_target]{\texttt{UpdateUnderstanding}} in Alg. \ref{algorithm: pseudocode}.
    \item \textit{Verify Student Bug Fixes (Section \ref{sec: restructuring}):} Each time the Student understands a target state variable, they are asked to provide, if any, recommended bug fixes based on the conversation history. This serves as an early stopping condition. This is \texttt{\hyperref[prompt: i2i_apply_bug_fixes]{\texttt{isResolved}}} in Alg. \ref{algorithm: pseudocode}.
\end{enumerate}


\subsection{State Space Estimation}
\label{sec: state space representation}
\par The goal of state space estimation is to determine the optimal criteria to track a Student's global understanding of a problem $P$ and their code, such that from the initial buggy state $B$, we can traverse the space to reach the goal state (correct code $C$).

\par We define the state space as the set of all possible tasks that a Student could perform to correct their buggy code. We claim that the optimal state space can be represented by a series $S$ of $k$ tasks which leads the Student from their buggy code $B$ to (1) understanding their conceptual and syntactical mistakes and (2) correcting their code. Each of these tasks is a \textbf{state variable} $\tau_i$ which either has a value of \textit{True} or \textit{False} based on whether the Student has completed it. At the very beginning of the Instructor-Student conversation, all of these variables are set to ``False''. We provide the estimated state space used in Figures \ref{fig:framework} and \ref{fig:tree}.
\begin{enumerate}[leftmargin=*]
\itemsep-0.25em 
    \item $\tau_1$: \textbf{False}, \textit{Understand the definition of the Fibonacci Sequence.}
    \item $\tau_2$: \textbf{False}, \textit{Recognize that the recursive call only returns the sequence till the ${(n-2)}^{\text{th}}$ term.}
    \item $\tau_3$: \textbf{False}, \textit{Modify the recursive call from fib-\\onacci(n-2) to fibonacci(n-1).}
\end{enumerate}
\par The state variables $\tau_i$ are structured such that earlier tasks have a higher priority, as their completion may consequently resolve later tasks. For instance, a student's buggy code may reflect that they do not conceptually understand the definition of the Fibonacci sequence. However, once this misunderstanding is resolved, the Student may simultaneously correct their related syntactical mistakes. On the other hand, attempting to resolve their syntactical mistakes, ``Modify the recursive call'', beforehand may lead to an unproductive and less structured conversation overall.

\subsection{Tree-Based Questioning}
\label{sec: tree-based questioning}
Tree-based questioning helps to structure the logical flow of the conversation and allows for more relevant, personalized questions. We use a tree to encode the Student's path to understanding at least one specific target state variable $\tau_i$. In each tree, (1) nodes are questions, (2) sibling nodes reflect questions which aim to \textit{sequentially} solidify the current misunderstanding, and (3) each of the parent-child edges connect nodes that guide to new understanding. Guided by the state space in Section \ref{sec: state space representation}, each level $l$ in the tree has questions $q$ of a similar difficulty and depth; the last level of the tree indicates that a specific state variable has been resolved. The Verifier agent dictates the movement from level to level and tree to tree.

\paragraph{Conditional generation of sibling questions.} The Instructor \textit{conditionally generates} sibling questions at level $l$ if and only if the \textit{Student incorrectly answers the Instructor question} (lines 6 and 10 in Alg. \ref{algorithm: pseudocode}). As shown in the second and third question of Figure \ref{fig:tree}, these questions must lead to the same level of target understanding as the original generated question intended therefore, the question can be rephrased or made more specific. To ensure this, we ground the question generation based on two things: (1) the previous questions from level $l$, and (2) the Verifier's explanation for why the Student got the question wrong.

\paragraph{Conditional generation of child questions.} The Instructor \textit{conditionally generates} child questions at level $l+1$ if and only if the \textit{Student correctly answers the Instructor question} (addresses the question and has no mistakes in their answer), but still does not understand the target state variable $\tau_i$ (line 14 in Alg. \ref{algorithm: pseudocode}). As shown in the fourth question of Figure \ref{fig:tree}, these questions aim to guide the Student to a more complete understanding of the target state. To ensure this, we ground the question generation on two things: (1) the previous questions from level $l-1$, and (2) the Verifier's explanation of the gaps in the Student's target state understanding.

\subsection{Adaptive Conversation Restructuring}
\label{sec: restructuring}

Once the Verifier agent determines that the target state/task has been resolved, we exploit the same process to update all remaining tasks $\tau \in S$, as multiple dependent bugs may have been concurrently resolved within the same tree. After at least the target state variable has been resolved (line 13 in Alg. \ref{algorithm: pseudocode}), we create a new tree for any remaining tasks, as shown in the first interaction of Figure \ref{fig:tree}. This step is crucial to the multi-bug setting, as independent bugs would benefit from having separate, distinct trees of questioning.

\par For further adaptiveness to the conversation, we additionally provide (1) an early stopping condition based on the Student's intermediate bug fixes, and (2) a maximum tree width and depth threshold, after which TreeInstruct chooses to teach the Student their remaining gap in knowledge.
\vspace{-2mm}
\noindent
\begin{itemize}[leftmargin=*]
    \item \textit{Bug fixes:} After a task $\tau$ has been resolved, the Student is prompted to provide a list of natural language bug fixes (e.g. ``Replace \texttt{i} with \texttt{i+1} on line 6.'') based on their entire conversation history with the Instructor. The Verifier will determine if all the ground-truth bug fixes have an \textit{isomorphic} counterpart within the set of suggested Student bug fixes. Isomorphism can be defined as (1) having the same conclusion or output, (2) sharing the same underlying logical structure or pattern, and/or (3) being convertible to each other through a series of logical transformations\footnote{For example, the bug fixes "Replace \texttt{if i <= 0} with \texttt{if i < 0}" and "Check if \texttt{i} is strictly less than 0" are isomorphic because they have the same conclusion and share the same underlying pattern of the mathematical operator $<$.}. If all ground-truth bug fixes have been resolved, then we may stop early.
    \item \textit{Teaching:} After generating a maximum number of sibling questions $q$ or depth $l$, the Instructor appends the correct answer to $Q[l][0]$ and re-ask $Q[l][-1]$ to the Student. This ensures that the conversation flows in case the Student gets stuck. 
\end{itemize}

\begin{table*}[ht!]
\center \small
\caption{Results on Socratic Debugging Benchmark (Single Bug). \textbf{Bolded} and $\dagger$ values denote the top 2 methods.}
\label{table: single_bug}
\begin{tabular}{lccccccccc}
\toprule
& \multicolumn{1}{c}{} & \multicolumn{4}{c}{Syntactical (42 samples)} & \multicolumn{4}{c}{Conceptual (107 samples)} \\
\cmidrule(lr){3-6} \cmidrule(lr){7-10}
Methods & \#T & Success & Relevant & Indirect & Logic & Success & Relevant & Indirect & Logic \\
\midrule
Vanilla \llama & 3.23 & \textbf{80.95} & 83.72$^\dagger$ & 76.19 & 78.70$^\dagger$ & 76.64$^\dagger$ & 87.35$^\dagger$ & 80.32$^\dagger$ & 78.79$^\dagger$ \\
Bridge \llama & 6.00 & 78.57$^\dagger$ & 76.50 & 82.24$^\dagger$ & 41.72 & 62.14 & 78.12 & 79.86 & 34.38 \\
\addlinespace
\hline
\addlinespace
\textbf{TreeInstruct \llama} & 5.41 & 77.27 & \textbf{92.01} & \textbf{96.48} & \textbf{88.95} & \textbf{80.26} & \textbf{95.63} & \textbf{89.10} & \textbf{94.63} \\
\bottomrule
\end{tabular}
\end{table*}

\section{Experiments}

\subsection{Experimental Setup}
In order to evaluate TreeInstruct, we utilize a proxy \textit{Student agent} based on the \texttt{Mistral-7B-Instruct} model  \cite{jiang2023mistral} to mimic the abilities of a student while responding to the Instructor. The prompt we use to define the Student persona is outlined in Table \ref{prompt: student_persona} of Appendix \ref{appendix: prompts}. We additionally provide GPT-4 API experimental setup details in Appendix \ref{sec: api_setup}. 

\subsection{Datasets}
\label{section: datasets}
We evaluate our method on two datasets. First, we use the Socratic Debugging Benchmark dataset from \cite{al2023socratic}, which consists of 149 problems-- each with a problem statement, student buggy code, bug fixes and descriptions in English, and correct code. However, these problems lack sufficient difficulty, often requiring small fixes and minimal problem comprehension. To evaluate TreeInstruct on more challenging problems, we craft a novel dataset, \textbf{MULTI-DEBUG}, based on 50 popular programming problems\footnote{https://github.com/Garvit244/LeetCode} (16 easy, 29 medium, and 5 hard according to their corresponding LeetCode labels). It features longer problems and more involved concepts, requiring TreeInstruct to have more extensive reasoning capabilities for guiding the Student. 

\par For each problem, expert annotators (Appendix \ref{appendix: evaluators}) injected 1, 2, and 3 syntactical or conceptual bug(s) that a typical student would make (a total of 150 different samples for MULTI-DEBUG). Conceptual bugs usually cause runtime errors or result in incorrect output. Examples include misunderstanding the problem statement, encountering an infinite loop, or incorrectly using a library or mathematical operator ($/$ vs $//$ in Python). Syntactical bugs cause compilation errors due to incorrect Python syntax (e.g., missing a colon). For each bug, we keep track of its fix and description.

\subsection{Baselines}
We compare TreeInstruct to several baselines and their variants. Given that LLM-based Socratic instruction is a new evolving area with few existing work and no multi-turn methods, we adapt a prompting and an existing single-turn method to our task. The first baseline, \textbf{Vanilla}, is given the same input as TreeInstruct's Instructor. The base model is prompted to ask Socratic questions to the Student-- it does not use any explicit conversational structure or estimate the Student's knowledge. We use both Meta-Llama-3-8B-Instruct \llama \cite{touvron2023llama} and GPT-4 \gpt \cite{achiam2023gpt} as base models for the Vanilla baseline.


\begin{table*}[ht!]
\center \small
\caption{Results on the \textbf{MULTI-DEBUG} dataset. In total, 1-bug has 29 syntactical and 21 conceptual bugs, 2-bug has 50 syntactical and 50 conceptual bugs, and 3-bug has 78 syntactical and 72 conceptual bugs. \textbf{Bolded} and $\dagger$ values denote the top 2 methods, respectively. The bottom two rows are ablation studies performed on 3-Bug setting.}
\label{table: multi_bug}
\begin{tabular}{lc|ccccccccc}
 \toprule
& \multicolumn{1}{c}{} & \multicolumn{1}{c}{} & \multicolumn{4}{c}{Syntactical} & \multicolumn{4}{c}{Conceptual} \\
\cmidrule(lr){4-7} \cmidrule(lr){8-11}
Bugs&Methods& \#T & Success & Relevant & Indirect & Logic & Success & Relevant & Indirect & Logic \\
\midrule
\multirow{4}{*}{1} & Vanilla \gpt        & 2.36       & 71.43          & 92.16      & 55.12      & 84.15  & \textbf{78.57}     & 94.58      & 59.17      & 84.17  \\
                   & BRIDGE \gpt         & 16.60       &  50.00         & 93.93      & \textbf{98.04 }     & 24.23  & 68.00      & 97.27      & \textbf{96.67}      & 35.38  \\
                   & \textbf{TreeInstruct} \llama & 7.24       & \textbf{76.19}     & 93.98$^\dagger$      & 94.08      & 85.28$^\dagger$  & 71.43     & 97.57$^\dagger$      & 93.02$^\dagger$      & 86.02$^\dagger$  \\
                   & \textbf{TreeInstruct} \gpt    & 3.94       &  75.00$^\dagger$         & \textbf{100.00}     & 95.59$^\dagger$      & \textbf{96.63}  & 76.92$^\dagger$         & \textbf{100.00}     & 88.01      & \textbf{94.76}      \\ 
      \addlinespace\hline\addlinespace

\multirow{4}{*}{2} & Vanilla \gpt        & 8.32       &  53.26         & 83.45      & 74.41      & 60.82  & 62.50          & 86.96      & 74.13      & 59.90  \\
                   & BRIDGE \gpt         & 15.28      & 34.88          & 89.47      & 89.33      & 52.40  & 42.71     & 89.67      & 88.06      & 46.64   \\
                   & \textbf{TreeInstruct} \llama & 9.04       & 66.67$^\dagger$        & 93.00$^\dagger$      & 92.17$^\dagger$      & 84.59$^\dagger$  & 72.62$^\dagger$          & 94.15$^\dagger$      & 92.58$^\dagger$      & 81.46$^\dagger$  \\
                   & \textbf{TreeInstruct} \gpt    & 6.14       & \textbf{69.32}          & \textbf{97.96}      & \textbf{98.47}      & \textbf{90.14}  &  \textbf{73.91}       & \textbf{99.58}      & \textbf{98.47}      & \textbf{94.45}      \\ 
      \addlinespace\hline\addlinespace

\multirow{6}{*}{3} & Vanilla \gpt        & 17.48      & 44.00$^\dagger$          & 69.88      & 64.31      & 52.38  & 67.00          & 84.68      & 84.68      & 41.51  \\
                   & BRIDGE \gpt         & 8.44       & 19.00         & 87.78      & 83.95      & 64.95  & 43.00     & 90.09      & 85.78      & 44.65  \\
                   & \textbf{TreeInstruct} \llama & 10.46      & 43.00         & 95.68$^\dagger$      & 88.88      & 80.94$^\dagger$  &  72.00$^\dagger$         & 96.76$^\dagger$      & \textbf{97.95}      & 83.28$^\dagger$  \\
                   & \textbf{TreeInstruct} \gpt    & 10.46      & \textbf{73.00}        & \textbf{100.00}     & \textbf{99.27}      & \textbf{95.57}  & \textbf{92.00}          & \textbf{98.40}      & 95.89$^\dagger$      & \textbf{93.63}      \\ 
                    \addlinespace\cline{2-11}\addlinespace
                   & \textbf{\llama No State}    & 16.34     & 25.51      & 51.61      & 97.21      & 41.09  & 53.00          & 47.57      & 94.70      & 20.36  \\ 
                   & \textbf{\llama No Teaching}    &  9.69     & 30.61      & 90.75      & 97.61$^\dagger$      & 72.84  & 50.00          & 94.62      & 95.17      & 68.78  \\ 
      \addlinespace\hline\addlinespace
\end{tabular}
\end{table*}

Second, we use \textbf{BRIDGE} \cite{wang2023bridging}. Since we are adapting this for Socratic code debugging, we use their predetermined error types, remediation strategies, and remediation intentions to guide the question generation, along with the problem-specific input given to TreeInstruct's Instructor. For both baselines, we limit the conversations to 20 turns per number of bugs.

\subsection{Evaluation Metrics:}
\label{sec: main_paper_metrics}
We perform qualitative and quantitative evaluation of our methods. The scores are averaged across all turns and then averaged across all problems. In the result tables, we scale the scores by 100.

\paragraph{Qualitative:} 
We develop a binary scale to assess the Socratic quality of questions. \cite{al2023socratic} identifies multiple dimensions of Socratic questioning, including relevance to specific bugs, implicitness of the answer, and structural coherence. For each question, we measure the below attributes of the conversation manually (giving a score of 1 if the attribute is met, and 0 otherwise):
\noindent
\begin{itemize}[leftmargin=*]
\itemsep-0.15em 
    \item \textbf{Relevance (Relevant):} The instructor's question was pertinent to the errors in the student's code.
    \item \textbf{Indirectness (Indirect):} The instructor's question refrained from directly revealing solutions to the bugs.
    \item \textbf{Logical Flow (Logic):} The instructor's question promoted a coherent conversation, facilitating the student's problem-solving process.
\end{itemize}

\paragraph{Quantitative:} We apply quantitative metrics to objectively evaluate the effectiveness and efficiency of our framework.
\noindent
\begin{itemize}[leftmargin=*]
\itemsep-0.15em 
    \item \textbf{Overall Success Rate (Success):} We check whether the final list of bug fixes generated by the Student, $B_S$, and the ground truth set of big fixes, $B_{GT}$, are isomorphic (Section \ref{sec: restructuring}). The success rate is calculated as $|B_{GT} \cup B_S| / |B_{GT}|$.
    \item \textbf{Average Number of Turns (\#T):} We compute the average number of turns required by the method to reach the goal state. This metric provides insight into the efficiency and depth of the interaction process.
\end{itemize}




\subsection{Overall Results}
In Tables \ref{table: single_bug} and \ref{table: multi_bug}, we see that with \textbf{more structured representations of student knowledge and conversation state}, TreeInstruct demonstrates \textbf{significant improvements beyond the baselines}. Across all multi-bug settings, we see an overall improvement of 16.6\% and 11.59\% in the success rates for syntactical and conceptual bugs, respectively. We also see an improvement of 13.47\% and 14.89\% for syntactical and conceptual bugs, respectively, across the three conversation metrics. For 1-bug, we see that the Vanilla \gpt baseline has the highest success for conceptual bugs, \textit{and} the lowest Indirectness score, indicating that questions were very direct, and gave hints towards the bug fixes, which evidently increased the success rate. We see the same trend in the syntactical, single bug on Vanilla \llama setting in Table \ref{table: single_bug}. Overall, TreeInstruct demonstrates \textbf{strong performance despite drastically different base models, \llama and \gpt.} TreeInstruct's runtime performance is detailed in Appendix \ref{appendix: runtime}.

\begin{table}[h!]
\center \small
\caption{Results on the side-by-side evaluation. \textbf{Bolded} and $\dagger$ values denote the top 2 comparisons, respectively. Note: S-bug refers to the Socratic Debugging Benchmark. We abbreviate TreeInstruct as TI.}
\label{table: side-by-side}
\begin{tabular}{l|cccc}
\toprule
Comparison                & 3-bug  & 2-bug & 1-bug & S-bug \\
\midrule
\addlinespace
TI vs BRIDGE  & 71.43$^\dagger$  & 68.00$^\dagger$ & \textbf{83.67} & \textbf{94.63} \\
TI vs Vanilla & \textbf{100.00} & \textbf{90.00} & 69.39$^\dagger$ & 50.33$^\dagger$ \\
BRIDGE vs Vanilla       & 57.14  & 62.00 & 40.82 & 24.83 \\
\addlinespace
\bottomrule
\end{tabular}
\end{table}

\paragraph{Side-by-Side Evaluation:} Using the same evaluation metrics (Section \ref{sec: main_paper_metrics}), we performed a side-by-side evaluation that measures how often a user prefers TreeInstruct over the baselines. The results are shown in Table \ref{table: side-by-side}, in which we see that on average, TreeInstruct is given a \textbf{higher ranking than BRIDGE 79.43\%} of the time, and a \textbf{higher ranking than Vanilla 77.43\%} of the time. The details of this computation are in Appendix \ref{appendix: side-by-side}. When specifically comparing TreeInstruct against BRIDGE, the key differences were that BRIDGE began with specific questions which revealed bug fixes and ended with general questions, lacking a logical flow in the conversation. In addition to being limited in a multi-bug setting, it posed less effective follow-up questions. In contrast, TreeInstruct prioritized high-level conceptual questions early on in order to build the foundation for later code-specific questions. Additionally, it asked for new information each turn and handled multiple bugs effectively. This is explicitly demonstrated in the example shown in Appendix \ref{appendix: interdependency}, Table \ref{table: bug_interdependency}.

\paragraph{Human Student Interaction:} We conducted a separate case study where human students directly interacted with TreeInstruct. We gathered five human volunteers of varying levels of programming backgrounds and knowledge, and presented each volunteer with three Single Bug problems and three 3-bug (MULTI-DEBUG) problems. The study shows that TreeInstruct can adapt to various levels of students effectively, as the scores are mostly comparable between our volunteers. The volunteers mentioned that \textbf{TreeInstruct helped them learn programming concepts by forcing them to critically think about their mistakes}, instead of \textit{trying random solutions}. Also, our volunteers had better interactions with TreeInstruct \gpt than TreeInstruct \llama because the \gpt Verifier is of higher quality. Appendix \ref{appendix: human_student} includes further details of this study, with quantitative results in Table \ref{table: human_student}.

\subsection{Ablation Studies}
\par We perform two ablations on all 50 3-bug problems (bottom two rows of Table \ref{table: multi_bug}). For \textbf{\llama No State}, we remove the state space representation, basing the question generation on the conversation history, Verifier feedback on the Student's answer, and the Student-proposed bug fixes. For \textbf{\llama No Teaching}, we remove the teaching functionality that kicks in after three consecutive incorrect Student answers. The conversation is still guided by the state space representation, tree-based questioning, its updates, and the Student-proposed bug fixes.

When compared to TreeInstruct \llama, \textbf{No State success rates drop by 18.25\%, Relevance by 46.63\%, and Logic by 51.39\%}. Without the grounding provided by the state space representation, the conversations deviate from the real bugs and contain many repeated questions that the Student already answered. Additionally, when compared to TreeInstruct \llama, \textbf{No Teaching success rates and Logic scores for drop by 17.20\% and 11.32\%, respectively}. When the Student does not know some base-level of foundational knowledge, it is better to break the unnecessary cycle of asking questions and allow the Instructor to teach these concepts. Overall, these performance differences demonstrate the impact of each component.

\paragraph{Verifier Subtask Performance:} To understand the Verifier's abilities to perform its designated subtasks (Section \ref{sec: agents}), given their complexity, we evaluate the Verifier on each subtask for the 3-bug setting. We demonstrate that for both \llama and \gpt, on the most difficult setting, both Verifier models feature high accuracy rates across all subtasks (86.3\% average performance across all subtasks), most notably \texttt{generateState}, which shows that the base model does not have significant impact. Overall, this shows that TreeInstruct can reliably depend on the Verifier to complete such tasks. The detailed results are included in Appendix \ref{appendix: subtasks}.

\subsubsection{Analysis}

\paragraph{Fine-grained and dependency-aware state spaces improve conversation quality.} 
In our evaluation, we saw that state spaces typically contain 4–5 state variables, evenly distributed across all bugs. In other words, the number of bugs is inversely proportional to the number of variables allocated to each bug. Thus, 1-bug uses all state variables-- each featuring a more fine-grained subtask to solve the bug, whereas 2- or 3-bug would use fewer variables for each bug (example provided in Appendix \ref{appendix: state space comp}). 1-bug's fine-grained state variables lead to increased Success and Relevance scores as the conversation delves deep into the Student's root misunderstanding of the bug. On the other hand, 3-bug features more coarse-grained state variables with a higher \textit{inter-bug dependency}. Consider the (recursive) Fibonacci problem in Figure \ref{fig:case_sg}. Suppose the student is missing a base case and incorrectly calling the recursive function. Solving one bug requires adequate understanding of the other, thereby making it easier to solve. Another example can be found in Appendix \ref{appendix: interdependency}, Table \ref{table: bug_interdependency}. The 2-bug setting does not \textit{fully} experience the benefits of fine-grained state variables or high bug-interdependency, resulting in a slight dip in scores: on average, a 7.62\%, 0.83\%, and 2.85\% drop in Success, Relevant, and Logic scores, respectively-- nonetheless, better than the baselines.

\paragraph{TreeInstruct tackles challenging problems more effectively.} Given the increase in problem difficulty between (1) the Socratic Debugging Benchmark (SDB) and MULTI-DEBUG (MD) datasets, and (2) from the 1 to 3-bug settings within MD, TreeInstruct is able to tackle challenging problems more effectively than baselines. The state space estimation breaks down problems into simple subtasks which the Instructor and Verifier can target using tree-guided questioning, even in complex conversations. Concretely, TreeInstruct's performance in the SDB setting (Table \ref{table: single_bug}) is comparable to that of the MD setting (Table \ref{table: multi_bug}), with only a 2.44\% drop in performance from SDB to MD 1-bug and 1.53\% drop from MD 1-bug to MD 3-bug. Comparatively, Vanilla \gpt drops by 14.55\% from 1-bug to 3-bug. Overall, TreeInstruct's Relevant, Indirect, and Logic scores remain high regardless of difficulty.

\section{Conclusion}
This paper proposes a novel method, \textbf{TreeInstruct}, for state space estimation and dynamic tree-based questioning for multi-turn Socratic instruction. We construct a novel multi-bug debugging dataset, \textbf{MULTI-DEBUG}, with 150 expert-annotated conceptual and syntactical problems and buggy solutions/fixes. Extensive experiments on an existing benchmark and MULTI-DEBUG demonstrate that TreeInstruct can be universally applied to both open and closed source models. We also showcase that TreeInstruct's strong Socratic questioning abilities widely outperform all baselines through both (1) rigorous quantitative and qualitative expert evaluation (preferred over 77.94\% of the time), and (2) real-world interactions with students of varying coding abilities.

\section{Limitations \& Future Work}
While TreeInstruct provides an intuitive framework which demonstrates promising results for effective multi-turn Socratic instruction, it contains a few limitations that form the foundation for future, impactful research areas.

Firstly, Tables \ref{table: single_bug} and \ref{table: multi_bug} shows high qualitative scores for the questions asked by TreeInstruct. While these are encouraging, the success rates still have large room for improvement-- the highest success rate is 77.27\%. This indicates that Socratic questions alone are not sufficient for teaching a student to debug their code. We judge the efficacy of questions locally, whereas the next step would be to judge them globally across the conversation. We leave it to future work to devise an effective global questioning scheme and evaluation metric.

Additionally, our method is dependent on the base model's reasoning capabilities, specifically for the Verifier agent. In our results, with a stronger model, we see higher scores for Logic and Success. Although our method shows comparable results between GPT-4 and Llama-3-8b, this may be a bottleneck, as stronger and bigger models require a higher deployment cost.

Next, in the few failure cases, we see some adverse effects of our method's reliance on the reasoning capabilities of the base model. First, our method can get stuck into a cyclical conversation with the Student if they are particularly weak in an area and cannot understand the target state even after multiple rounds of direct questioning and teaching. In these cases, the number of turns rises to 20-30.

Moreover, syntactical bugs might be ``harder to see'' for the language model, as it goes against the generation process to generate syntactically incorrect code. Breaking it down, a language model is trained to generate code with a colon at the end of for loops, if-statements, and method signatures, so if buggy code has a missing colon, the language model might ignore it. This results in syntactical bugs being harder to solve.

These limitations give way to exciting future work. Firstly, we can make use of vision language models to provide students with multi-modal teaching strategies, instead of relying solely on language. Additionally, we can enhance the framework, so it will explore new instruction methods when the questioning becomes cyclical. This can also help make the Instructor more reliable to generate consistent output across multiple runs on the same problem. Furthermore, we can utilize a structured fine-tuning approach to help the model better leverage the Verifier feedback and tree-based question generation process to make hierarchical Socratic planning and questioning inherent to a model. Overall, TreeInstruct can also be extended to automatically generalize to different teaching domains (e.g., quantitative reasoning).

\section{Ethics Statement}

We are committed to the transparency and reproducibility of our research. We encourage our research community to make use of our open-source code and dataset to further improve our methodology. Our research involves the evolving intersection of large language models (LLMs) and education, where the deployment of language model instructors and their interactions with students have been relatively unexplored. The role of technology and language models is being widely discussed with respect to its impact on student dependence and lack of critical thinking. Given the rapid and wide-scale deployment of LMs to the public, we emphasize the importance of designing Socratic dialogue systems in the hopes of bettering educational support for all students and educators.

\section{Acknowledgements}

\par This research project has benefited from the Microsoft Accelerate Foundation Models Research (AFMR) grant program, through which leading foundation models hosted by Microsoft Azure and access to Azure credits were provided to conduct the research. Furthermore, we would like to thank Mihir Kavishwar, Krish Agarwal, Sonia Agarwal, Nirav Diwan, and Shradha Sehgal for their help and feedback on our work.

\bibliography{anthology,custom}
\bibliographystyle{acl_natbib}


\appendix
\begin{table*}[ht!]
\center \small
\caption{A comparison of questions asked by BRIDGE and TreeInstruct - between Q1 and Q2, and Q2 and Q3, the Student gave a correct answer.}
\label{table: bug_interdependency}
\begin{tabular}{l|p{0.45\textwidth}|p{0.45\textwidth}}
\toprule
 & \textbf{BRIDGE}  & \textbf{TreeInstruct} \\
\midrule
\addlinespace
Q1         & What is the difference between a list and a set in Python, and how might this affect the performance of your code when checking if an item is in 'temp'? & Can you explain the difference between a list and a set in Python, and how their properties might affect the functionality of your code?                     \\
Q2         & Can you explain why you chose to use a list to store the characters of the string and what might be the implications of this choice?                     & That is correct! Let's take a step further. Given that a set only contains unique items, how might this property of a set be useful in solving this problem? \\
Q3         & N/A                                                                                                                                                      & Can you explain how you are currently adding elements to your 'temp' collection in your code?                                                                \\

\addlinespace
\bottomrule
\end{tabular}
\end{table*}

\section{Breakdown of the Verifier Agent's Subtask Performance}
\label{appendix: subtasks}

We perform ablation studies for each module on our most difficult setting (MULTI-DEBUG 3-bug), with both GPT-4 and Llama-3-8B Verifier Agents. For the sake of time, we used 20\% of the dataset. For each subtask (\texttt{generateState}, \texttt{verifyResponse}, \texttt{updateUnderstanding}, \texttt{isResolved}), we define the following metrics of success:

\begin{itemize}
    \item\texttt{\textbf{generateState}}: the proportion of bugs that would be sufficiently resolved if the tasks (state space variables) were followed by the Student.
    \item \texttt{\textbf{verifyResponse}}: a binary 0/1 for correct Verifier judgment and 0/1 for the correct explanation behind the Verifier's judgment: $\frac{\text{verification\_score} + \text{explanation\_score}}{2}$.
    \item \texttt{\textbf{updateUnderstanding}}: the proportion of state variables that were correctly validated (whether to update the state variable) and 0/1 for the correct explanation behind the Verifier's judgment: $\frac{\text{validation\_score} + \text{explanation\_score}}{2}$.
    \item \texttt{\textbf{isResolved}}: the proportion of Student proposed bug fixes which the Verifier correctly identifies as isomorphic/non-isomorphic to the corresponding ground truth bug fixes.
\end{itemize}

The results are in Table \ref{table: verifier_impact}. Despite this being our most difficult setting, both Verifier models feature high accuracy rates across all subtasks, most notably \texttt{generateState}, which shows that the choice of base model does not have significant impact. Note: \texttt{isResolved} is primarily used for early stopping (e.g., if a student proposes all correct bugs during the first turn, then TreeInstruct stops early), so this subtask is not necessary. This subtask simply reduces computational costs.

\begin{table}[h]
\center\small
\caption{Performance of each Verifier subtask.}
\label{table: verifier_impact}
\begin{tabular}{l|cc}
\toprule
Subtasks & Llama-3-8B  & GPT-4 \\
\midrule
\addlinespace
\texttt{\textbf{generateState}}       & 90.0       & 100.0 \\
\texttt{\textbf{verifyResponse}}      & 85.7       & 86.7  \\
\texttt{\textbf{updateUnderstanding}} & 88.6       & 85.9  \\
\texttt{\textbf{isResolved}}          & 72.2       & 81.4  \\
\addlinespace
\bottomrule
\end{tabular}
\end{table}

\section{Code Bug Interdependency}
\label{appendix: interdependency}

Table \ref{table: bug_interdependency} shows an example of the effect of interdependent bugs, taken from real conversations that BRIDGE and TreeInstruct had with our student simulator \texttt{Mistral-7B-Instruct}. The questions aim to solve: (1) using a set instead of a list, and (2) using \texttt{.add()} instead of \texttt{.append()} to add items. As shown, Questions 1 and 2 tackle Bug 1, and Question 3 tackles Bug 2. The interdependency of changing the data structure first makes it easier to resolve the second bug, yielding higher success rates for the 3-bug setting.

\begin{table*}[th!]
\caption{Inter-annotator agreement on the relevance, indirectness, logic scores, and bug type determination. Conducted on a random subset of the Single-Bug (SB) and MULTI-DEBUG 3-Bug (3-Bug) datasets.}
\label{tab: inter-annotator}
\center \small
    \begin{tabular}{c|cccccccccc}
 \toprule
\multicolumn{5}{c}{Cohen's $\kappa$} & \multicolumn{4}{c}{\% Agreement} \\
\cmidrule(lr){2-5} \cmidrule(lr){6-9}
Setting  & Success & Relevant & Indirect & Logic & Success & Relevant & Indirect & Logic \\
\midrule
        \textbf{SB} & 61.97 & 71.88 & 76.62 & 1.0 & 91.67 & 94.44 & 88.89 & 1.0 \\
        \textbf{3-Bug} & 1.0 & 78.90 & 78.90 & 1.0 & 1.0 & 97.83 & 97.83 & 1.0 \\
      \addlinespace\hline\addlinespace

\end{tabular}
\end{table*}

\input{app_side_by_side}

\section{Human Expert Evaluators}
\label{appendix: evaluators}
As mentioned before, the injected bugs into the dataset (Section \ref{section: datasets}) and evaluation in Tables \ref{table: single_bug} and \ref{table: multi_bug} were obtained using volunteer human expert evaluators: two computer science teaching assistants with at least four years of high-school, undergraduate, and graduate-level teaching experience, with proficiency in Python and located in the USA. Their experiences equipped them with the first-hard experience of mistakes made by beginner programmers, leading to more realistic injected bugs. Furthermore, for evaluation, they were given the same set of instructions and the following set of guidelines:
\begin{itemize}
    \item Assign a score of 1 for Relevance if the question will eventually lead the Student to understand their bug(s).
    \item Assign a score of 0 for Indirect if a question explicitly or implicitly states a solution.
    \item Assign a score of 0 for Logic if the current question does not naturally flow from the Student's previous answer.
\end{itemize}

They were given a CLI-based annotation system. In our GitHub repository, we have provided the script (\href{https://github.com/agarwalishika/TreeInstruct/blob/main/evaluation/question_human_evaluation.py}{question\_human\_evaluation.py}) and an example of the evaluation interaction (\href{https://github.com/agarwalishika/TreeInstruct/blob/main/evaluation/evaluation_example.pdf}{evaluation\_example.pdf}). Below are some special cases/considerations the evaluators were also given:
\begin{itemize}    
    \item If the Verifier is wrong and asks the same question despite the Student getting the question correct, give a score of 0 for Relevance.
    \item If a question seems out of order, give a score of 0 for Logic.
    \item If a question deep into the conversation is vague, gives a score of 0 for Relevance and Logic.
    \item If the answer is provided in a hint after 2 rounds of similar questions, and the Student still does not understand, do not penalize the Instructor for Indirect.
    \item For determining Success, do not penalize the Student if the bug fix is in natural language rather than code.
\end{itemize}

\par On a random subset of the Single-Bug and MULTI-DEBUG 3-Bug datasets, we computed the inter-annotator agreement on the relevance, indirectness, and logic scores (Section \ref{sec: main_paper_metrics}) and bug type determination. We show these agreement results in Table \ref{tab: inter-annotator}.

\input{app_state_space_comp}

\begin{table*}[ht]
\center \small
\caption{Results of human student evaluation across S(ingle)-bug (Socratic Debugging benchmark) and 3-bug (\textbf{MULTI-DEBUG} dataset) settings, broken down by the student level (Lvl).}
\label{table: human_student}
\begin{tabular}{lc|ccccccccc}
 \toprule
& \multicolumn{1}{c}{} & \multicolumn{1}{c}{} & \multicolumn{4}{c}{Syntactical} & \multicolumn{4}{c}{Conceptual} \\
\cmidrule(lr){4-7} \cmidrule(lr){8-11}
Bugs & Lvl & Avg. Turns & Success & Relevant & Indirect & Logic & Success & Relevant & Indirect & Logic \\
\midrule
\multirow{5}{*}{S-bug \llama} & 1  & 6.0 & \textbf{100.00} & 66.67 & 66.67 & \textbf{100.00}& \textbf{100.00} &91.67$^\dagger$ & \textbf{100.00}& 50.79   \\
                               & 2  & 12.0& \textbf{100.00} & 66.67 & 83.33$^\dagger$ & 75.00$^\dagger$ & 50.00$^\dagger$ & \textbf{100.00} & \textbf{100.00}& 50.00   \\
                               &  3  & 8.0 & 0.00$^\dagger$   & 87.50$^\dagger$ & \textbf{100.00}& 50.00 & \textbf{100.00} & 67.50 & 90.00$^\dagger$ & 42.50  \\
                               &  4  & 1.0 & \textbf{100.00} & \textbf{100.00}& 0.00  & \textbf{100.00}& \textbf{100.00} &57.14 & \textbf{100.00}& 64.29$^\dagger$ \\
                               &  5  & 1.0 & \textbf{100.00} & \textbf{100.00}& \textbf{100.00}& \textbf{100.00}& \textbf{100.00}& \textbf{100.00} & 75.00 & \textbf{75.00 } \\
      \addlinespace\hline\addlinespace

\multirow{5}{*}{3-bug \gpt} &  1  & 19.0 & 83.33$^\dagger$ & 75.93 & 97.92$^\dagger$ & 74.77& \textbf{100.00}& \textbf{100.00}& 88.89$^\dagger$ & 79.49   \\
                       &  2  & 11.7 & 83.33$^\dagger$ & \textbf{100.00}& \textbf{100.00}& 78.57& \textbf{100.00}& \textbf{100.00}& 86.67 & 82.50   \\
                       &  3  & 6.67 & \textbf{100.00}& \textbf{100.00}& \textbf{100.00}& 85.71$^\dagger$& \textbf{100.00}& \textbf{100.00}& \textbf{100.00}& \textbf{100.00}   \\
                       &  4  & 4.7  & \textbf{100.00}& 93.33$^\dagger$ & \textbf{100.00}& 76.67& \textbf{100.00}& \textbf{100.00}& 83.33 & 88.89$^\dagger$  \\
                       &  5  & 3.0  & \textbf{100.00}& \textbf{100.00}& 83.33&\textbf{100.00}& \textbf{100.00}& 83.33$^\dagger$ & \textbf{100.00}& 83.33  \\
      \addlinespace\hline\addlinespace
\end{tabular}
\end{table*}

\input{app_human_student}

\section{Model Inference Experimental Setup}
\label{sec: api_setup}
\subsection{GPT-4 API}
For GPT-4, we made use of OpenAI's GPT-4 API. Overall, we use temperature sensitivity $t=0$ for all generation tasks, except for $t=0.1$ for state space estimation and $t=0.3$ for instructor question generation. Using \$30 / 1M input tokens and \$60 / 1M output tokens, we break down the cost for each method. TreeInstruct uses an average of 35,000 input tokens and 4,000 output tokens, which adds up to \$\textbf{1.29 per conversation}. BRIDGE uses an average of 18,000 input and 5,500 output tokens, which adds up to \$\textbf{0.87 per conversation}. Vanilla uses an average of 31,000 output and 2,200 output tokens, which adds up to \$\textbf{1.06 per conversation}.

\subsection{Mistral and Llama}
We run the Mistral-7B-Instruct-0.2 and Llama models locally on 2 NVIDIA-RTX A6000 GPUs. For one pass on a dataset (i.e., 150 problems/conversations), TreeInstruct takes approximately 4 hours.

\subsection{Runtime Comparison}
\label{appendix: runtime}
We present the per-turn runtime for TreeInstruct (TI) across all multi-bug settings in Table \ref{table: runtime}. Llama was run on 1 NVIDIA Tesla V100 32GB, and OpenAI’s GPT-4 API was used for serverless inference. While the table indicates that Llama takes approximately 1 min per turn, this is due to a hardware bottleneck. With better hardware, TI-Llama should be able to run much faster to provide close to real-time tutoring. This is indicated by TI-GPT4 taking approximately 7 seconds per turn.

\begin{table}[ht]
\center
\caption{Average runtime (in seconds) TreeInstruct takes for each turn, for each setting.}
\label{table: runtime}
\begin{tabular}{l|c|c}
\toprule
\textbf{Setting}                & \textbf{Llama-3-8B}  & \textbf{GPT-4} \\
\midrule
\addlinespace
1-bug  & 56.27 & 7.09 \\
2-bug  & 65.69 & 7.77 \\
3-bug  & 61.22 & 7.65 \\
\addlinespace
\bottomrule
\end{tabular}
\end{table}

\section{License}
All the datasets used in this work, including our own, are under the Apache 2.0 License. Our use of existing artifact(s) is consistent with their intended use, specifically for the Socratic Debugging benchmark and in general, programming practice and feedback for the problems used in the MULTI-DEBUG dataset.

\input{prompts}

\end{document}

%% file: algorithm.tex
\begin{algorithm*}[h!]
\small
    \caption{TreeInstruct}
    \label{algorithm: pseudocode}
    \begin{algorithmic}[1]  
        \Require $P$ (Problem Description), $B$ (Buggy Code, Bug Descriptions), $C$ (Corrected $B$ Code, Bug Fixes)
        \State $S = \{\tau_1, \tau_2, \ldots, \tau_{k}\} \leftarrow \text{\hyperref[prompt: v2v_get_state_repr]{\text{GenerateState}}}(P, B, C)$ \Comment{Section \ref{sec: state space representation}: State representation: (resolved?, task)}
        \State $l \leftarrow 0$, $Q \leftarrow \{l:[]\}$, $H \leftarrow []$, $F \leftarrow \{\}$ \Comment{Tree level, question list/level, conv. history, Student bug fixes}
        \State $q \leftarrow \text{\hyperref[prompt: i2i_generate_initial]{GenerateQuestion}}(P, B, C, \tau_1)$ \Comment{Section \ref{sec: tree-based questioning}: Generate initial question }
        \break
        \While {$\exists \text{ } \tau \in S \ \text{s.t.} \  \neg\text{\hyperref[prompt: i2i_apply_bug_fixes]{\text{isResolved}} }(S, F, C)$} \Comment{Section \ref{sec: restructuring}: Process while tasks or bugs remain}
                \State $r \leftarrow \text{StudentResponse}(q)$
                \State $v, w \leftarrow \text{\hyperref[prompt: i2i_correct_response]{\text{VerifyResponse}}}(q,r)$ \Comment{Section \ref{sec: tree-based questioning}: is $r$ to $q$ correct ($v$); why or why not ($w$)?}
                \State $H.add(q, r)$
                \State $Q[l].add(q)$
                \If {$v = \text{false}$} \Comment{\textbf{Incorrect student response}}
                    \State $q \leftarrow \text{\hyperref[prompt: i2i_generate_sibling]{GenerateSiblingQuestion}}(\tau, Q[l], H, w)$ \Comment{Section \ref{sec: tree-based questioning}: factor in \textit{why} the student was incorrect}
                \Else \Comment{\textbf{Correct student response}}
                    \State $S, w \leftarrow \text{\hyperref[prompt: i2i_address_target]{\text{UpdateUnderstanding}}}(S, q, r)$ \Comment{Section \ref{sec: tree-based questioning}: tasks $\tau_i \ldots \tau_{k}$ resolved? If $\neg\text{S}[\tau]$, why ($w$)?}
                    \If {$\neg\text{S}[\tau]$}
                        \State $q \leftarrow \text{\hyperref[prompt: i2i_generate_child]{GenerateChildQuestion}}(\tau, Q[l], H, w)$ \Comment{Section \ref{sec: tree-based questioning}: factor in \textit{why} $\tau$ was unresolved}
                        \State $l \leftarrow l + 1$ \Comment{Advance to next tree level}
                        
                    \Else \Comment{\textbf{Task $\tau$ resolved}}
                        \State $F \leftarrow \text{\hyperref[prompt: i2s_generate_bug_fixes]{GetStudentBugFixes}}(H)$ \Comment{Section \ref{sec: restructuring}: ask Student for bug fixes (if any)}
                        \State $l \leftarrow 0, Q \leftarrow \{l:[]\}$ \Comment{$\tau$ resolved $\rightarrow$ create new tree}
                        
                    \EndIf
                \EndIf
        \EndWhile
    \end{algorithmic}
\end{algorithm*}

%% file: app_side_by_side.tex
\section{Side by Side Evaluation}
\label{appendix: side-by-side}

As mentioned in the main text, we perform a side-by-side evaluation to measure the percentage of times a user prefers our method TreeInstruct over the baselines. Preference was measured as the average of all conversation metrics across syntactical and conceptual bugs. Based on the metrics, we assign each method a ranking (1, 2, or 3). Table \ref{table: side-by-side} shows that TreeInstruct was preferred 68-94.6\% of the time over the baselines. On average, TreeInstruct was preferred over BRIDGE 79.43\% of the time, and over Vanilla 77.43\% of the time.

\paragraph{Interpretation.} When we say TreeInstruct was preferred 79.43\% more over BRIDGE, this means that across all 50 3-bug problems and ranking configurations, TI was given a higher ranking than BRIDGE (TI is ranked \#1 while BRIDGE is ranked \#2, TI is ranked \#1 while BRIDGE is ranked \#3, TI is ranked \#2 while BRIDGE is ranked \#3) 79.44\% of the time. Each of the 50 problems can have multiple preferences (TI over BRIDGE, TI over Vanilla, Bridge over Vanilla, etc.) which is why they will not necessarily add up to 100.

Table \ref{table: bug_interdependency} shows a comparison between the questions asked by BRIDGE and TreeInstruct. Appendix \ref{appendix: interdependency} specifies the details behind the targeted problem in the Table. On the left, we see that BRIDGE asks very similar questions regarding the design choice of using a list versus a set. TreeInstruct, on the right, asks one question regarding the design choice, one question about how to add elements to the correct data structure, and one intermediate question about why the design choice is suited towards the problem. Updating the state space, grounding questions towards the Student's capabilities, and connecting concepts back to the problem are principles that rank TreeInstruct higher than BRIDGE.

%% file: app_state_space_comp.tex
\section{Comparing State Space Representations in Multi-Bug Settings}
\label{appendix: state space comp}
Here, we compare the state space representations of the 1-bug, 2-bug, and 3-bug settings for the two sum problem. In the two sum problem, given is an array of integers and a target value. The goal is to return the indices of two numbers that add up to the target value. Below is the correct code.

\begin{lstlisting}
1. def twoSum(self, nums, target): 
2.   d = {}
3.   for i in range(len(nums)):
4.     difference = target-nums[i]
5.     if difference in d:
6.       return [d[difference], i]
7.     d[nums[i]] = i
8.   return d
\end{lstlisting}

In the 1-bug setting, the Student mistakenly writes \verb|nums[i]-target| instead of \verb|target-nums[i]| on line 4. In the 2-bug setting, along with the previous bug, the Student also initializes \verb|d| as a list (\verb|d=[]|) instead of a dictionary on line 2. Finally, in the 3-bug setting, the Student forgets to add a colon at the end of the if-statement on line 5. 

Tables \ref{ssc: state_one}, \ref{ssc: state_two}, and \ref{ssc: state_three} outline the state space representations for the 1-bug, 2-bug, and 3-bug settings. As shown, 1-bug uses 3 states (states 1, 2, and 3) to solve the same but that 3-bug uses 1 state (state 1) to solve. This means the 1-bug state representation is much less compact than that for 3-bug.

\begin{table}[h]
\centering
\begin{tabularx}{\columnwidth}{|X|}
\hline
1. Understand the problem statement and the requirement to find two numbers that add up to a specific target. \\
2. Understand the logic behind calculating the difference as target - nums[i]. \\
3. Correctly implement the difference calculation in the code.
 \\ \hline
\end{tabularx}
\caption{State space representation for 1-bug on the two-sum problem.}
\label{ssc: state_one}
\end{table}

\begin{table}[h]
\centering
\begin{tabularx}{\columnwidth}{|X|}
\hline
1. Understand how to correctly calculate the difference between the target and the current number in the array. \\
2. Understand the difference between lists and dictionaries in Python. \\
3. Correctly initialize a dictionary in Python. \\
4. Understand how to use a dictionary to store and retrieve values in Python.
 \\ \hline
\end{tabularx}
\caption{State space representation for 2-bug on the two-sum problem.}
\label{ssc: state_two}
\end{table}

\begin{table}[h]
\centering
\begin{tabularx}{\columnwidth}{|X|}
\hline
1. Understand how to correctly calculate the difference as `target-nums[i]`. \\
2. Understand how to initialize a dictionary using `{}` instead of `[]`. \\
3. Understand how to use a dictionary to store and retrieve values. \\
4. Understand the correct syntax for an if-condition, including the necessary colon at the end.
 \\ \hline
\end{tabularx}
\caption{State space representation for 3-bug on the two-sum problem.}
\label{ssc: state_three}
\end{table}

%% file: app_human_student.tex
\section{Interactive Evaluation with Human Students}
\label{appendix: human_student}

For our main evaluation, we used \texttt{Mistral-7B-Instruct} to represent a Student. We noticed that Mistral is an overconfident model that (1) suggests incorrect bug fixes in between the conversations and (2) jumps to fix bugs that do not exist in the code. Therefore, we worked with human students to test our method on the following two settings: Socratic Debugging on TreeInstruct \llama and 3-bug on TreeInstruct \gpt. We gathered 5 human volunteers of varying levels of programming backgrounds and knowledge (ensuring to anonymize their identities):

\begin{itemize}
    \item \textbf{Level 1:} Student knows how TreeInstruct works; they act as an adversary to intentionally provide bad inputs that will try to make the method fail.
    \item \textbf{Level 2:} Student is new to TreeInstruct; they are a basic programmer who has been learning to code in Python for a few months.
    \item \textbf{Level 3:} Student is new to TreeInstruct; they are a non-computer science major who does not use Python often, but knows the basic high level concepts of data structures and syntax.
    \item \textbf{Level 4:} Student is new to TreeInstruct; they have been using Python for 2 years and are in their final year of undergraduate education in computer science.
    \item \textbf{Level 5:} Student knows how TreeInstruct works; they act as an ally to intentionally provide good inputs, so the method can resolve the bugs in as little turns as possible.
\end{itemize}

\noindent When conducting the study, we adhered to the following experimental process:
\begin{enumerate}
    \item We presented the student with the problem statement and gave them as much time as they needed to fully understand it.
    \item The students were given two minutes to review the buggy code. We noted down how many bugs each of the students were able to identify before their conversation.
    \item The students conversed with TreeInstruct until they were able to identify all the bugs present in the code.
\end{enumerate}

We provide the results of this interactive study in Table \ref{table: human_student}. We used the same three single and 3-bug questions for all students, leading to 30 human student interactions in total. We also conducted a post-interaction interview with each of the students and provide an overview of their feedback below:
\newline\newline
\noindent\textbf{Socratic questioning helped students learn programming concepts.} The Level 3 student stated that, \textit{"If there was no conversation, I would be put off from attempting to fix and just try a bunch of different things based on the errors."} Overall, students of Levels 2-4 (students with no knowledge of the system) were not able to identify all the bugs before their interactions, but ended up solving them independently under the Socratic guidance of TreeInstruct.
\newline\newline
\noindent\textbf{Underlying model had a significant impact on user experience.} Students had a significantly better experience with TreeInstruct \gpt compared to TreeInstruct \llama. Specifically, the quality of the Verifier determined whether the questions posed by the Instructor would be overly repetitive or not.

\subsection{Analysis}
Table \ref{table: human_student} contains the results. We see that from Level 1 to Level 5, the conversation have fewer turns, especially in the 3-bug setting. Additionally, we see that syntactical bugs are harder to solve for weaker students (on average, a success rate of 86.67\%), which is intuitive as these students do not have a strong foundation in Python syntax. On the other hand, conceptual bugs are easier to solve (on average, a 95\% success rate). Overall, the results show that our method can adapt to various levels of students effectively.

%% file: prompts.tex
\section{Prompts}
\label{appendix: prompts}
A few of the prompts use one-shot learning, and the fields are prefixed with "example". These examples are hand chosen, with no criteria in mind. The example problem relates to a solution that outputs the Fibonacci sequence of length $n$, where $n$ is the input. We provide the specific prompts starting from the next page.

\begin{table*}[h]
\centering
\begin{tabularx}{\textwidth}{|X|}
\hline
You are an Instructor helping a Student debug their code to solve the following problem statement (after tag 'problem'). You have access to their buggy code (after tag 'bug\_code'). Do not ask questions that explicitly or implicitly mention the following:
 \\ \hline
\end{tabularx}
\caption{Instructor agent persona prompt}
\label{prompt: instructor_persona}
\end{table*}

\begin{table*}[]
\centering
\begin{tabularx}{\textwidth}{|X|}
\hline
You are a Student writing code to solve the above problem statement (after tag 'problem'), and you have written the below buggy code (after tag 'buggy\_code'). You are seeking help from your Instructor help solve your 'buggy\_code'. Your role is to answer the questions that the Instructor asks you as if you were an introductory programmer with a beginner's level of coding knowledge.
 \\ \hline
\end{tabularx}
\caption{Student agent persona prompt}
\label{prompt: student_persona}
\end{table*}

\begin{table*}[]
\centering
\begin{tabularx}{\textwidth}{|X|}
\hline
You are an assistant to the Instructor helping a Student debug their code to solve the following problem statement (after tag 'problem'). Your role is to determine the Student's understanding (or lack thereof) within the Instructor-Student interactions. You have access to the correct code (after tag 'correct\_code'). Assume the Student is a introductory programmer with a beginner's level of coding knowledge.
 \\ \hline
\end{tabularx}
\caption{Verifier agent persona prompt}
\label{prompt: verifier_persona}
\end{table*}

\begin{table*}[]
\centering
\begin{tabularx}{\textwidth}{|X|}
\hline

Given the student's buggy code (after tag 'buggy\_code'), bug description (after tag 'bug\_description'), bug fixes (after tag 'bug\_fixes'), and the correct code (after tag 'correct\_code') for solving the problem statement (after tag 'problem'), we define the state representation of a set of Instructor-Student interactions as a series of necessary tasks which lead the Student from their 'buggy\_code', with bugs described in 'bug\_description', to understanding and correcting their conceptual and syntactical mistakes to reach 'correct\_code' with the 'bug\_fixes'. \\
\\
We define a state representation as a list of state attributes, where each attribute denotes a specific task that is NECESSARY for the student to successfully understand and implement the given problem. A NECESSARY task directly addresses at least one of the 'bug\_description's and thus, is NOT ALREADY ADDRESSED in 'buggy\_code'. In other words, if a task is not successfully completed, the Student will never be able to correct their 'buggy\_code' to 'correct\_code'. \\
\\
If the student's 'buggy\_code' shows that they have already understood and implemented a specific task, DO NOT INCLUDE that task as a state attribute since it is REDUNDANT. \\
\\
The list should be ordered, with earlier attributes/tasks given priority over later ones (e.g., conceptual understanding tasks are a pre-requisite and thus more important than syntactical tasks). The following is an example of the state representation for the given example problem statement: example problem: Implement a Fibonacci sequence using recursion. \\
--- \\
\texttt{In-context example from Table \ref{prompt: state_repr_icl}} \\
---

In the above example format ('example\_explanation' and 'example\_state\_representation'), output an 'explanation' for your plan on which state\_attributes to output and the 'state\_representation' with all NECESSARY 'state\_attribute's for the following 'problem' statement based on 'correct\_code', the Student's 'buggy\_code', the 'bug\_description's, and its respective 'bug\_fixes':

\\ \hline
\end{tabularx}
\caption{Internal Verifier prompt to estimate the state space representation; corresponds to the GenerateState() method in line 1 of Alg \ref{algorithm: pseudocode}.}
\label{prompt: v2v_get_state_repr}
\end{table*}

\begin{table*}[]
\centering
\begin{tabularx}{\textwidth}{|X|}
\hline
Input: \\
example\_problem: \\
Implement a fibonacci sequence using recursion. 
\\\\
example\_buggy\_code:

    def Fibonacci(n):
        if n <= 0:
            print("Incorrect input")
        elif n == 1:
            return 1
        else:
            return Fibonacci(n-1)
\\\\
example\_bug\_description: \\
On line 7, the function only recursively calls `Fibonacci(n-1)`, which will then only return `1` from the edge case on line 5. Instead, the function should consider that the nth term of the Fibonacci sequence is computed as the sum of the preceding n-1 and n-2 values.
\\\\
example\_bug\_fixes: \\
Replace `return Fibonacci(n-1)` with `return Fibonacci(n-1) + Fibonacci(n-2)` on line 7.
\\\\
example\_correct\_code: \\

    def Fibonacci(n):
        if n <= 0:
            print("Incorrect input")
        elif n == 1:
            return 1
        else:
            return Fibonacci(n-1) + Fibonacci(n-2)
\\\\
Output:

example\_explanation: Based on the 'bug\_description', the bug involves an incorrect recursive call, which indicates that the Student does not understand that the Fibonacci sequence requires taking the sum of the preceding two terms for getting the current value. An example where `n` is equal to `2` points this mistake out. Finally, based on the 'bug\_fixes', the Student must modify the recursive call to add `Fibonacci(n-2)` in order to reach the `correct\_code` state.
\\\\
example\_state\_representation:\\
state\_attribute\_1: Understand the definition of the Fibonnaci Sequence.\\
state\_attribute\_2: Consider the preceding two items in the sequence for computing the current value in the Fibonacci Sequence.\\
state\_attribute\_3: Consider the example when `n` is equal to `2`.\\
state\_attribute\_4: Correctly recursively call `Fibonacci(n-2)` and add it to the existing `Fibonacci(n-1)`.\\
 \\ \hline
\end{tabularx}
\caption{In context example for Table \ref{prompt: v2v_get_state_repr}}
\label{prompt: state_repr_icl}
\end{table*}

\begin{table*}[]
\centering
\begin{tabularx}{\textwidth}{|X|}
\hline
The Student has written code (after tag 'student\_code') to solve the problem (after tag 'problem') and is answering a question (after tag 'Student') from the Instructor (after tag 'Instructor') based on their understanding of the 'problem' and their 'student\_code'. IF the Student suggests a solution to a bug they identify, also consider the following: \\
\\
Ensure that the Student's suggestion is isomorphic to any one of the bug fixes mentioned in the provided 'bug\_fixes'; if not, then 'answer\_has\_no\_mistakes' should be "False". A Student's suggestion is isomorphic to a bug fix if they (1) have the same conclusion or output, (2) share the same underlying logical structure or pattern, and (3) are convertible to each other through a series of logical transformations. \\
\\
Answer the following questions and within your reasoning, think about how you would answer the "instructor\_question" yourself and include this in your "explanation".:
answer\_addresses\_question: <Does the Student's response (after tag 'Student') directly answer the Instructor's question (after tag 'Instructor')? Output "True or "False">
answer\_has\_no\_mistakes: <Is the Student's response (after tag 'Student') to the Instructor's question (after tag 'Instructor') logical (no logical errors or mistakes)? Output "True or "False"> \\
\\
Instructor: \{Instructor question\} \\
Student: \{Student response\} \\
bug\_fixes: \{bug fixes\} \\
student\_code: \{student code\} \\
\hline
\end{tabularx}
\caption{Internal Verifier prompt to assess the accuracy of the Student response with respect to the Instructor's question; corresponds to the VerifyResponse() method in line 6 of Alg \ref{algorithm: pseudocode}.}
\label{prompt: i2i_correct_response}
\end{table*}

\begin{table*}[]
\centering
\begin{tabularx}{\textwidth}{|X|}
\hline
A Student has sufficient understanding of a certain topic (specified at tag "target\_understanding") when the responses that they provide to the Instructor (specified in the "conversation\_history") would REQUIRE them to comprehend "target\_understanding". This can either be demonstrated (1) explicitly, where the Student directly mentions "target\_understanding", OR (2) implicitly, where their reasoning is isomorphic to completing the task in "target\_understanding". A Student's reasoning is isomorphic to the "target\_understanding" if they (1) have the same conclusion or output, (2) share the same underlying logical structure or pattern, and (3) are convertible to each other through a series of logical transformations. \\
\\
Based on the Student's response (after tag 'student\_response') to the Instructor's question (after tag 'instructor\_question') and the conversation history (after tag 'conversation\_history'), do you believe that the Student needed to sufficiently comprehend the "target\_understanding" in order to provide their responses (after tag 'Student' in 'conversation\_history') to the Instructor's questions (after tag 'Instructor' in 'conversation\_history') throughout the conversation history? Include specific quotes from the "conversation\_history" in your "explanation". Within your reasoning, think about how you would answer the "instructor\_question" yourself and include this in your "explanation". \\
\\
Instructor: \{Instructor question\} \\
Student: \{Student response\} \\
target\_understanding: \{target understanding\} \\
\hline
\end{tabularx}
\caption{Internal Verifier prompt to update the state space with respect to a single-turn Instructor-Student interaction; corresponds to the UpdateUnderstanding() method in line 12 of Alg \ref{algorithm: pseudocode}.}
\label{prompt: i2i_address_target}
\end{table*}

\begin{table*}[]
\centering
\begin{tabularx}{\textwidth}{|X|}
\hline
Are any bug fixes mentioned in the conversation that you have had with the Instructor (under tag "conversation\_history")? If no, return "None". If yes, then follow the format below: \\
\\
First, based on your current understanding of the problem (tag "problem") and your conversation with the Instructor, summarize (after tag "bug\_summarization") the bugs in the code explicitly mentioned within the "conversation\_history" that you believe will revise your buggy code (after tag "buggy\_code") to a correct implementation of the "problem" statement. Then, based on this summary, output a list of the explicitly mentioned bug fixes (from "bug\_fix\_1" to "bug\_fix\_n", where $n$ is the number of bug fixes to make), each described briefly. \\
\\
An example format/wording of a brief bug fix would be: "Replace `i` with `i+1` on line 6." \\
\\
conversation history: \{convo history\} \\
 \\ \hline
\end{tabularx}
\caption{Instructor to Student prompt that asks the Student to generate a list of bug fixes; corresponds to the GetStudentBugFixes() method in line 17 of Alg \ref{algorithm: pseudocode}.}
\label{prompt: i2s_generate_bug_fixes}
\end{table*}

\begin{table*}[]
\centering
\begin{tabularx}{\textwidth}{|X|}
\hline
For the problem description given above (after tag 'problem'), you are given two sets of bug fixes (under tags 'suggested\_bug\_fixes' and 'correct\_bug\_fixes'). For each bug fix in 'correct\_bug\_fixes', is there at least one bug fix in 'suggested\_bug\_fixes' that is isomorphic? Two bug fixes are isomorphic if they (1) have the same conclusion or output, (2) share the same underlying logical structure or pattern, and (3) are convertible to each other through a series of logical transformations. Output "True" or "False" as your answer with an explanation. \\
\\
suggested bug fixes: \{student\_bf\} \\
\\
correct bug fixes: \{correct\_bf\} \\
 \\ \hline
\end{tabularx}
\caption{Internal Verifier prompt check if the Student has suggested all the correct bug fixes that are present in the ground truth set of bug fixes, corresponds to isResolved() in line 4 of Alg. \ref{algorithm: pseudocode}.}
\label{prompt: i2i_apply_bug_fixes}
\end{table*}

\begin{table*}[]
\centering
\begin{tabularx}{\textwidth}{|X|}
\hline
Based on the student's current level of understanding, as demonstrated through their conversation history (tag "conversation\_history"), what is 1 follow-up question with the same level of depth and difficulty RELATIVE to the 'previous\_questions' that you could ask based on the Student's explanation that would help them reach the "target\_understanding"? Make sure that the question addresses the reasons why the Student got the previous question(s) wrong, as detailed in tag "misunderstanding", such that the Student is more likely to resolve these misunderstandings.
You must generate a question such that any correct answer to your question should automatically reflect the "target\_understanding" and resolve the "misunderstanding".\\
\\
target\_understanding: \{target\}\\
conversation\_history: \{conversation history\} \\
previous\_questions: \{previous questions\}\\
previous\_misunderstanding: \{explanations\} \\

These questions should help the Student arrive at the answer themselves; do NOT give any direct hints towards the solution (under tag "bug\_fixes" and tag "bug\_description").

bug\_fixes: \{bug fixes\} \\
bug\_descriptions: \{bug descriptions\} \\
\hline
\end{tabularx}
\caption{Internal Instructor prompt to generate a sibling question; corresponds to the GenerateSiblingQuestion() method in line 10 of Alg \ref{algorithm: pseudocode}.}
\label{prompt: i2i_generate_sibling}
\end{table*}

\begin{table*}[]
\centering
\begin{tabularx}{\textwidth}{|X|}
\hline
Based on the student's current level of understanding, as demonstrated through their conversation history (tag "conversation\_history"), what is 1 follow-up question with increasing depth and difficulty RELATIVE to the 'previous\_questions' that you could ask based on the Student's explanation that would help them reach the "target\_understanding"? Make sure that the question addresses the reasons why the Student has not reached the "target\_understanding", as detailed in tag "misunderstanding", such that the Student is more likely to resolve these "misunderstanding"s by answering your question. \\
\\
target\_understanding: \{target\}\\
conversation\_history: \{conversation history\} \\
previous\_questions: \{previous questions\}\\
previous\_misunderstanding: \{explanations\} \\

These questions should help the Student arrive at the answer themselves; do NOT give any direct hints towards the solution (under tag "bug\_fixes" and tag "bug\_description"). \\
\\
bug\_fixes: \{bug fixes\} \\
bug\_descriptions: \{bug descriptions\} \\
\hline
\end{tabularx}
\caption{Internal Instructor prompt to generate a child question; corresponds to the GenerateChildQuestion() method in line 14 of Alg \ref{algorithm: pseudocode}.}
\label{prompt: i2i_generate_child}
\end{table*}

\begin{table*}[]
\centering
\begin{tabularx}{\textwidth}{|X|}
\hline
Based on the buggy code and the target understanding state (under tag "target\_understanding"), what is one question (k=1) that you could ask that would help the Student reach the "target\_understanding"? 
These questions should help the Student arrive at the answer themselves; do NOT give any direct hints towards the solution (after tag 'bug\_fixes'). \\
\\
These questions should help the Student arrive at the answer themselves; do NOT give any direct hints towards the solution (under tag "bug\_fixes" and tag "bug\_description").\\
\\
target\_understanding: \{target\}\\
bug\_fixes: \{bug fixes\} \\
bug\_descriptions: \{bug descriptions\} \\
\hline
\end{tabularx}
\caption{Internal Instructor prompt to generate the initial question; corresponds to the GenerateQuestion() method in line 3 of Alg \ref{algorithm: pseudocode}.}
\label{prompt: i2i_generate_initial}
\end{table*}